\documentclass[10pt,twocolumn,letterpaper]{article}

\usepackage{cvpr}              

\usepackage[dvipsnames]{xcolor}

\definecolor{cvprblue}{rgb}{0.21,0.49,0.74}
\usepackage[pagebackref,breaklinks,colorlinks,citecolor=cvprblue]{hyperref}

\usepackage{bbm}

\usepackage{gensymb}
\usepackage{algorithm, algpseudocode}

\DeclareMathOperator*{\argmin}{arg\,min}

\def\paperID{14509} 
\def\confName{CVPR}
\def\confYear{2024}

\usepackage[accsupp]{axessibility} 
\usepackage{siunitx}
\usepackage{bm}

\title{MS-MANO: Enabling Hand Pose Tracking with Biomechanical Constraints}

\author{
Pengfei Xie\textsuperscript{1,*}, 
Wenqiang Xu\textsuperscript{2,*}, 
Tutian Tang\textsuperscript{2}, 
Zhenjun Yu\textsuperscript{2}, 
Cewu Lu\textsuperscript{2} \\
\textsuperscript{1}Southeast University
\textsuperscript{2}Shanghai Jiao Tong University\\
{ \small
\textsuperscript{1}{\tt\small xiepf2002@gmail.com}
\textsuperscript{2}{\tt\small\{vinjohn, tttang, jeffson-yu, lucewu\}@sjtu.edu.cn}
}\\
{
\small
\url{https://ms-mano.robotflow.ai}
}
}

\begin{document}
\maketitle
\renewcommand{\thefootnote}{}
\footnotetext{* Equal contribution.}

\begin{abstract}
This work proposes a novel learning framework for visual hand dynamics analysis that takes into account the physiological aspects of hand motion. The existing models, which are simplified joint-actuated systems, often produce unnatural motions. To address this, we integrate a musculoskeletal system with a learnable parametric hand model, MANO, to create a new model, MS-MANO. This model emulates the dynamics of muscles and tendons to drive the skeletal system, imposing physiologically realistic constraints on the resulting torque trajectories. We further propose a simulation-in-the-loop pose refinement framework, BioPR, that refines the initial estimated pose through a multi-layer perceptron (MLP) network. Our evaluation of the accuracy of MS-MANO and the efficacy of the BioPR is conducted in two separate parts. The accuracy of MS-MANO is compared with MyoSuite, while the efficacy of BioPR is benchmarked against two large-scale public datasets and two recent state-of-the-art methods. The results demonstrate that our approach consistently improves the baseline methods both quantitatively and qualitatively. 
\end{abstract}
\section{Introduction}
\label{sec:intro}

\begin{figure}
    \centering
    \begin{subfigure}[b]{\linewidth}
        \centering
        \includegraphics[width=1\linewidth]{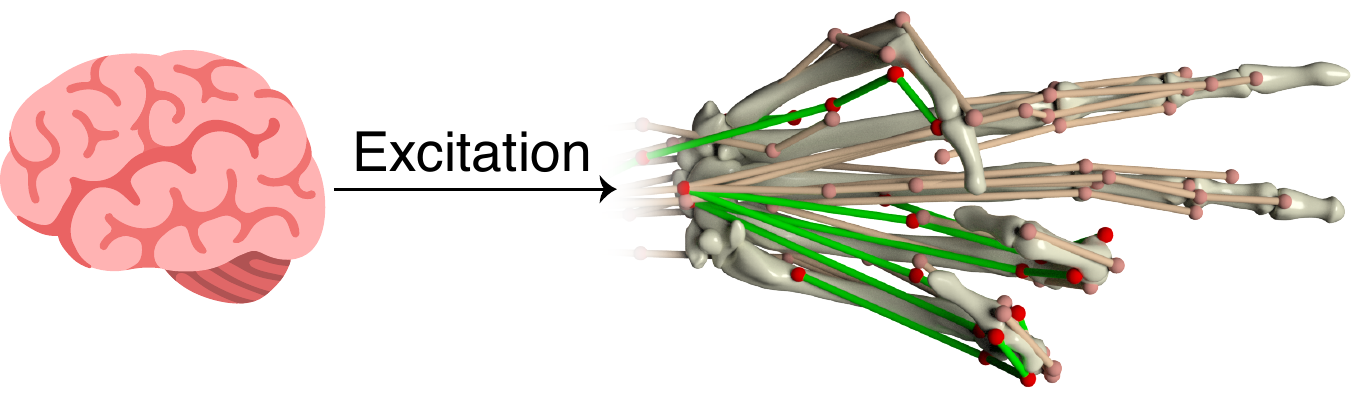}
        \caption{}
        \label{fig:fig1-a}
    \end{subfigure}
    \begin{subfigure}[b]{\linewidth}
        \centering
        \includegraphics[width=1\linewidth]{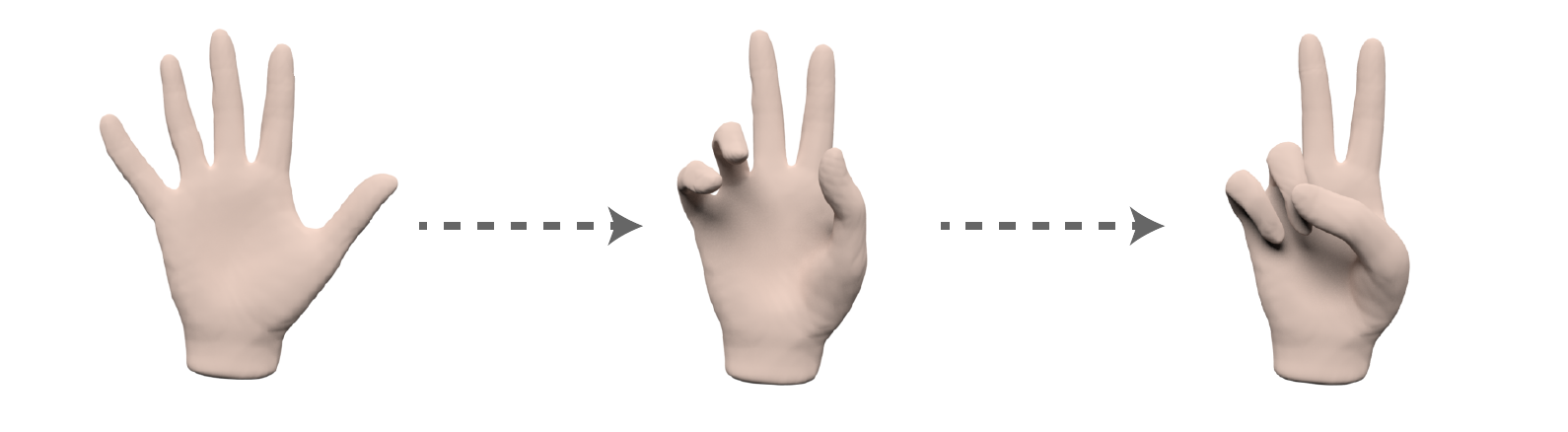}
        \caption{}
        \label{fig:fig1-b}
    \end{subfigure}
    \caption{The physiological mechanism of hand dynamics. \textbf{(a)} The excitation signal originating from the brain triggers the contraction or relaxation of muscles. The triggered muscle segments are illustrated in green, while the relaxed ones are in brown. \textbf{(b)} The muscle contraction triggered by excitation manifests as the movement of the hand in appearance.}
    \label{fig:musculoskeletal_system}
\end{figure}

From a physical perspective, human hand motion is actuated by the musculoskeletal system. As depicted in \cref{fig:musculoskeletal_system}, the brain transmits excitation signals via the nervous system, intriguing the contraction and relaxation in muscles and generating torque to facilitate joint movement of hands. Consequently, the dynamics of hand motion are naturally coordinated and constrained by the underlying musculoskeletal system. However, such physiological aspects are seldom taken into consideration when designing a learning framework of visual hand dynamics analysis (\eg hand pose estimation \cite{handtailor,pose1,bihand} and tracking \cite{gsdf,deformer}).

Previous works on visual hand analysis primarily considered hand dynamics as multi-body dynamics. This means that the hand is represented as an articulated object, with kinematic movement directly propelled by joint torques.
Since the joint-actuated system simplifies the mechanism of hand motion, it may produce robot-like movements that are unnatural or infeasible to human hands \cite{muscle_joint}. In contrast, a musculoskeletal system explicitly emulates the dynamics of muscles and tendons to drive the skeletal system so that it can impose physiologically realistic constraints on the resulting torque trajectories and make the movements more human-like. Despite the advantages, this system is challenging to replicate due to the complexity of the hand's dynamic system, which uses over 30 muscles to support nuanced movements.
In this work, we first integrate a musculoskeletal system with a parametric hand model, MANO \cite{mano}, extending it to the musculoskeletal version, \textbf{MS-MANO}. Then, we apply MS-MANO to the hand pose tracking task with a simulation-in-the-loop learning framework, \textbf{BioPR}.

To build MS-MANO, we focus on three key features: anatomical accuracy, support for learning tasks, and adaptivity to body shape variations. With accurate modeling, it can support precise control of subtle movements and achieves human-like motion; with the support of learning tasks, it can be integrated seamlessly into learning frameworks; with adaptivity, it can accommodate individuals with diverse body shapes.
To meet all these goals, we take the muscle-tendon data from MyoHand model \cite{myosuite}, which is built upon two anatomic data from OpenSim \cite{opensim}: MoBL model \cite{mobl1,mobl2} and the 2nd-Hand models \cite{2ndhand}. Then, we integrate this muscle-tendon data into the MANO model \cite{mano}, a widely recognized and adaptable hand model that can adjust to various hand shapes through different parameters.
We aim to have the muscle-tendon structure adaptable to a variety of shapes as well. To realize this, the bone-centric muscle representation is transformed into a joint-centric representation.
The details are discussed in \cref{sec:muscle_mapping}.

To show the utility of MS-MANO in visual learning tasks, we take the hand pose tracking as our experimental platform. Typically, spatial-temporal features are extracted from observed images. However, when the hand in the image is occluded or motion-blurred, these features may consequently be affected, leading to inconsistent prediction.
With biomechanical constraints provided by MS-MANO, we can largely mitigate such instability. To leverage MS-MANO in the hand pose tracking task, we propose a biomechanical pose refinement framework, \textbf{BioPR}. BioPR takes the predicted hand pose and velocity (which is inferred from multi-frame poses) as input. It first predicts the excitation signals for all the muscles in the hand model and then uses a simulator to run the hand motion with the excitation to get a reference pose. Next, BioPR refines the initial estimated pose by taking the estimated pose and the reference pose into a multi-layer perception (MLP) network.

The evaluation is conducted in two separate parts: the accuracy of MS-MANO and the efficacy of BioPR. To evaluate the accuracy of MS-MANO, we compare it with MyoSuite by calculating the difference in the trajectories generated. To evaluate BioPR, we adopt two large-scale public datasets that support hand pose tracking: DexYCB \cite{dexycb} and OakInk \cite{oakink} as benchmarks. Two recent state-of-the-art methods gSDF \cite{gsdf} and Deformer \cite{deformer} are selected as the baseline methods. With BioPR, the baseline methods are consistently improved quantitatively and qualitatively.

We summarize our contributions as follows:
\begin{itemize}[left=1pc]
    \item We present a musculoskeletal MANO (MS-MANO) hand model. It inherits all the merits of MANO, such as support of learning tasks and adaptivity to different body shapes, but also extends it with musculoskeletal modeling, which can ensure the biomechanical constraints for hand learning tasks.
    \item We exhibit the ability of MS-MANO in the hand pose tracking task with a simulation-in-the-loop framework, BioPR. We compare the performance of our method with multiple baseline methods on two different benchmarks.
\end{itemize}

\section{Related Works}

\subsection{Hand Dynamics System}

The dynamics of hand motion can be modeled in two distinct ways: through multi-body dynamics and biomechanics.
The former conceptualize the human hand as an articulated object, actuating hand movement by directly generating joint torques and may include biomechanical constraints \cite{Cai_2018_ECCV}. On the other hand, the latter approach creates musculoskeletal models that utilize biomimetic muscles and tendons to propel skeletal motion. Unlike joint-actuation models, muscle-actuation leads to movements that adhere to physiological constraints \cite{create_retarget}, and display energy expenditure more akin to actual humans \cite{muscle_energy}.
In the realm of computer animation, the control of a muscle-based virtual character has been investigated in relation to upper body movements \cite{upper1,upper2,upper3}, hand manipulation \cite{hand_manip1, hand_manip2}, and locomotion \cite{mass, muscle_energy,locomotion1,locomotion2,locomotion3,locomotion4}.

In recent years, the development of statistical parametric human body models has emerged \cite{mano,smpl,smplx}. A series of works have aimed to incorporate the musculoskeletal system into such models, enabling the musculoskeletal structure to adapt according to the parameters. BASH \cite{bash} integrates a musculoskeleton into the SMPL model. However, it lacks a complete muscle for the hand. Meanwhile, Ye et al. \cite{rcare} model a full-body musculoskeletal system, integrating it into the SMPLX model and modeling the mobility-limited behaviors for care recipients. However, these full-body systems do not meticulously model the hand muscle.

In biomechanics, hand movements are not just controlled by the hand but also by the muscles in the forearm. Therefore, a thorough musculoskeletal model for the hand should also include the forearm and wrist. MyoSuite \cite{myosuite} successfully integrates these into a single MyoHand model, which amalgamates anatomic data from the MoBL, a human upper extremity model \cite{mobl1,mobl2}, and the 2nd-Hand for hand and fingers models \cite{2ndhand}.
In line with BASH \cite{bash} and RCareWorld \cite{rcare}, we incorporate the MyoHand model into a parametric MANO model, resulting in our proposed MS-MANO model.

\subsection{Visual Hand Dynamics Analysis}
Analyzing hand dynamics visually typically involves estimating hand pose from a single image. The considerable advancement of learning-based research in this area is largely due to the parametric hand model, MANO \cite{mano}, and the differentiable layer that enables direct learning of hand parameters and generation of the hand model. Although the extraction of hand pose or kinematic structures from static images has achieved significant success, these methods \cite{pose1,cpf2,cpf1,artiboost,handtailor,h2o} are inherently incapable of predicting dynamic information.
Recently, some studies \cite{track1,arctic,gsdf,deformer,track2,track3,track4,track5,track6} have begun to investigate the temporal information in videos in order to regularize per-frame prediction. One approach explicitly models temporal information using techniques such as optical flow \cite{track1}, temporal consistency constraints \cite{track5, pose3}, and graph modeling \cite{graph_track}.
Another approach implicitly models temporal information by incorporating learning techniques with recurrent neural networks \cite{track2} or transformers \cite{deformer,gsdf}.
gSDF \cite{gsdf} adopts a signed distance field for both hand and object geometry and extracts the hand model with marching cube algorithm \cite{marching_cubes}. Then, the extracted hand mesh can be fitted to the MANO model to obtain the joint parameters.
Deformer \cite{deformer} adopts different transformer modules to extract spatial and temporal information. The features for each image are first extracted separately and then fused with a cross-attention, which improves the accuracy of hand pose estimation.

Different from previous works, instead of extracting spatial-temporal information from image observations, we provide a musculoskeletal prior and design a learning framework that uses this prior to adhere to biomechanical constraints.
\section{Musculoskeletal MANO, MS-MANO}
\begin{figure}
    \centering
    \begin{subfigure}[b]{\linewidth}
        \centering
        \includegraphics[width=\linewidth]{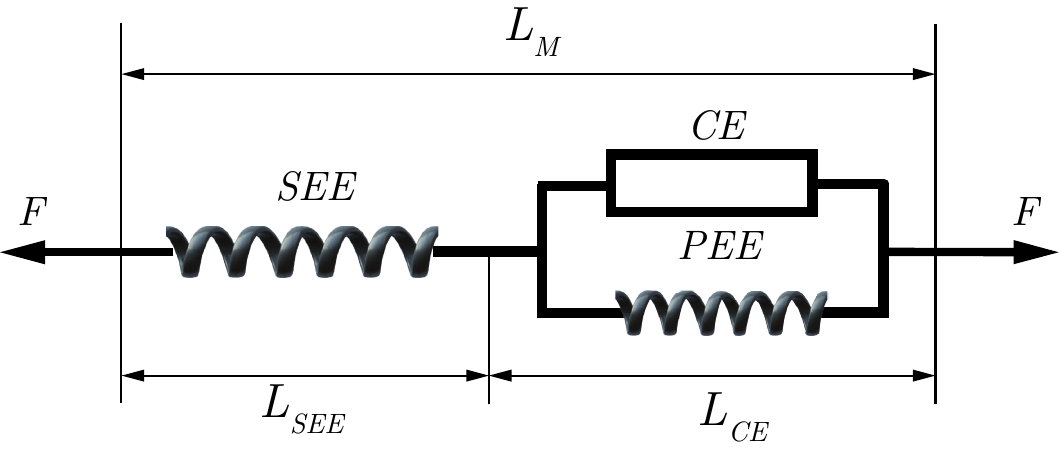}
        \caption{}
        \label{fig:hill-type-a}
    \end{subfigure}

    \begin{subfigure}[b]{\linewidth}
        \centering
        \includegraphics[width=\linewidth]{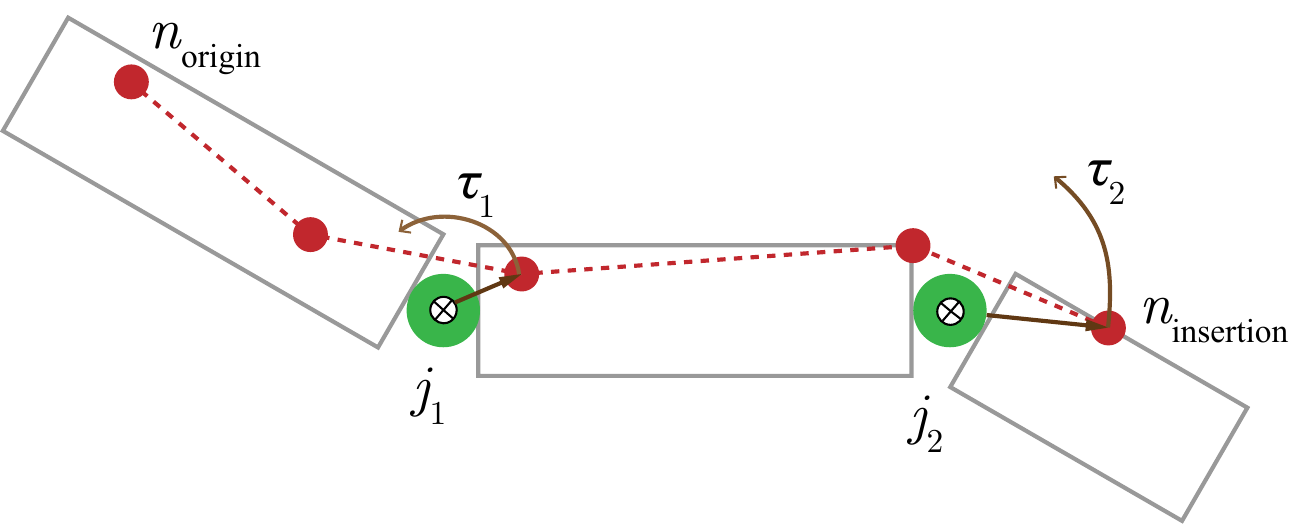}
        \caption{}
        \label{fig:hill-type-b}
    \end{subfigure}

    \caption{The hill-type muscle. \textbf{(a)} Each muscle segment is composed of the contractile element \textit{CE}, the parallel elastic element \textit{PEE}, and the serial elastic element \textit{SEE}. \textbf{(b)} Each muscle segment originates from a certain point $n_\text{origin}$ and ends at $n_\text{insertion}$. A joint $j$ connects two bones. Once triggered, the muscle segment can apply torque $\bm\tau$ on the joint.}
    \label{fig:hill_type}
\end{figure}
In this section, we will describe the musculoskeletal MANO (MS-MANO) model. We first give a brief introduction to the muscle model in \cref{sec:hill_type}. Then, we describe the muscle adaptation from bone-centric MyoHand data to joint-centric representation in \cref{sec:muscle_mapping}.

\subsection{Hill-type Muscle Model}\label{sec:hill_type}
Muscles are soft tissues that can generate forces to facilitate joint movements. To model the muscle dynamics behavior, we adopt the Hill-type model \cite{hilltype}, which is widely used in biomechanics \cite{opensim}. In the hill-type model (see \cref{fig:hill_type}), a muscle consists of segments represented by red dashed lines. Each line segment is modeled by three elements: the contractile element \textit{CE}, the parallel elastic element \textit{PEE}, and the serial elastic element \textit{SEE}. Each muscle initiates from a specific point $n_{\text{origin}}$ and triggers the muscle fiber, while the insertion points $n_{\text{insertion}}$ act as the remote endpoints and apply torque to the joint.

Thus, the torque for a joint can be calculated by:
\begin{equation}
    \bm\tau_\text{m} = f(F, x)\left\lVert(\bm q - \bm j)\times \frac{\bm s_c}{\lVert\bm s_c\rVert}\right\rVert,
    \label{eq:torque}
\end{equation}
where $F$ is the contractile force, $x$ is the muscle state, $\bm q$ is the point where the muscle is attached to the bone, $\bm j$ is the joint to apply torque on, and $\bm s_c$ is the muscle segment. The detailed deduction can be referred to \cite{rcare}.

\begin{figure*}
    \centering

    \begin{subfigure}[b]{0.3\linewidth}
        \centering
        \includegraphics[width=1\linewidth]{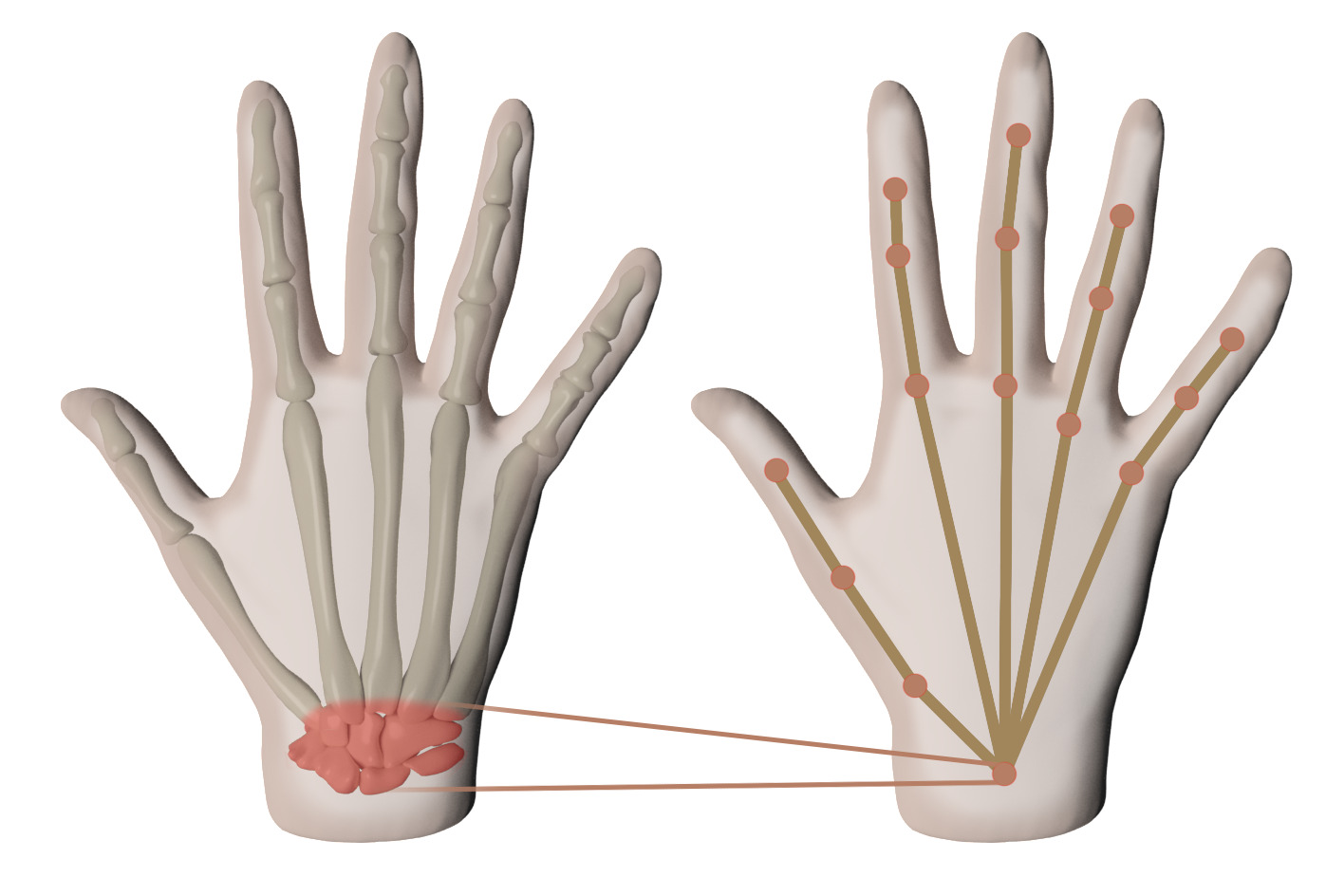}
        \caption{}
        \label{fig:mapping-a}
    \end{subfigure}
    \hfill
    \begin{subfigure}[b]{0.3\linewidth}
        \centering
        \includegraphics[width=1\linewidth]{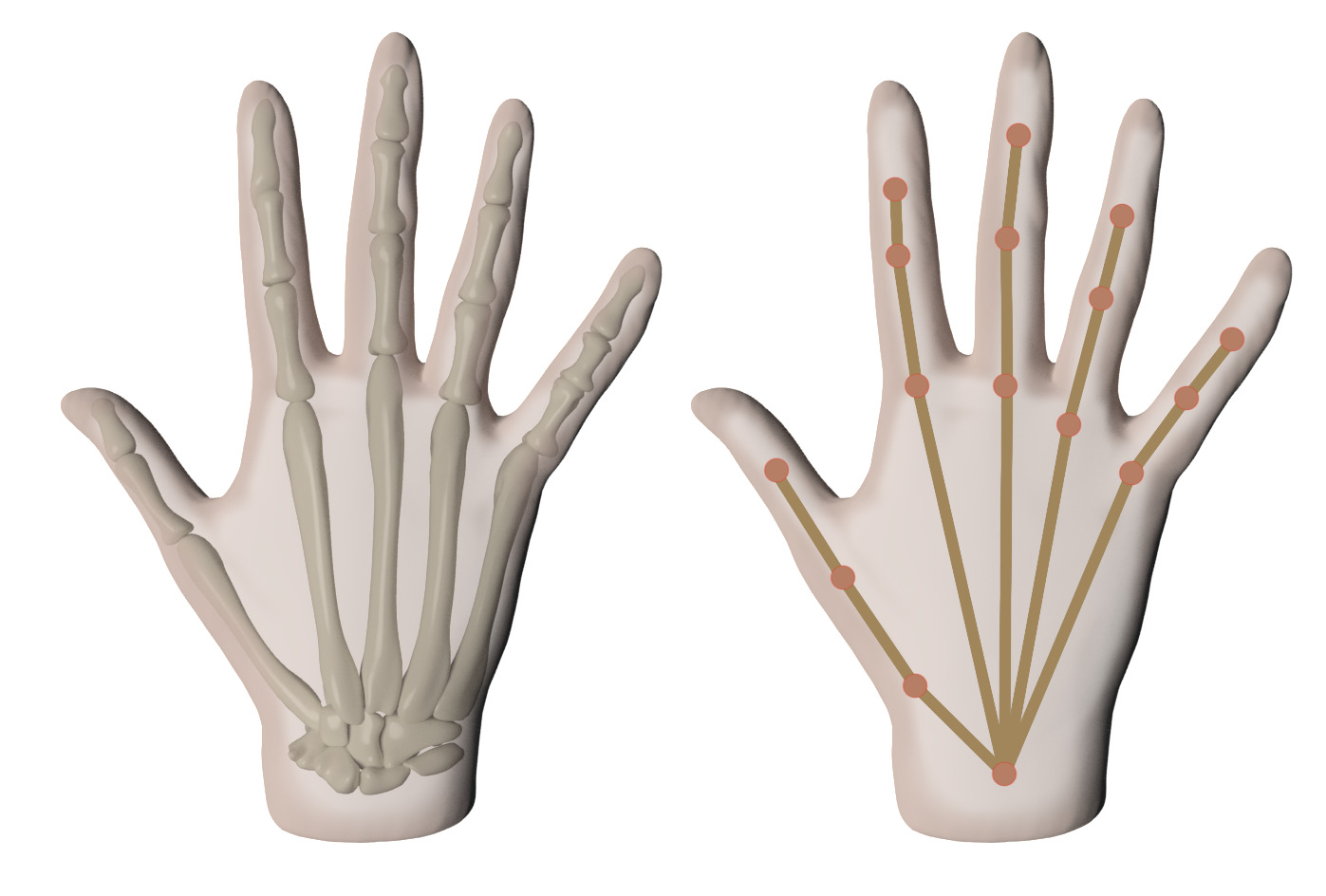}
        \caption{}
        \label{fig:mapping-b}
    \end{subfigure}
    \hfill
    \begin{subfigure}[b]{0.3\linewidth}
        \centering
        \includegraphics[width=1\linewidth]{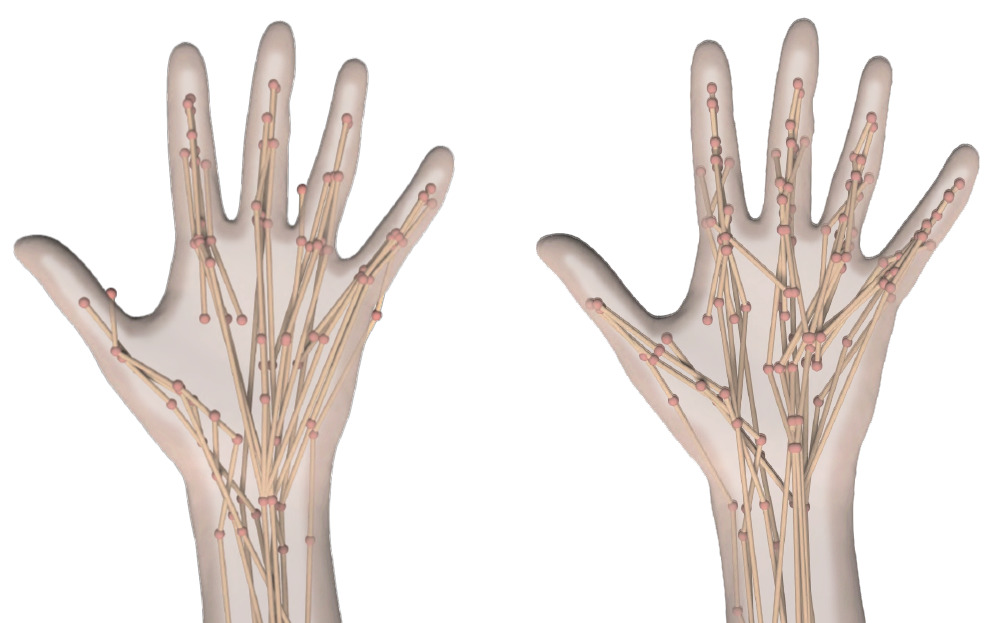}
        \caption{}
        \label{fig:mapping-c}
    \end{subfigure}

    \caption{Joint-centric muscle adaptation. \textbf{(a)} A set of smaller bones in the MyoHand model is mapped into a single joint in the MANO model. \textbf{(b)} The bone-centric muscle segments can adapt to different shapes. \textbf{(c)} (Left) The raw skeleton after the automatic mapping will result in issues like intersection. (Right) The manually revised skeleton can perfectly fit with the MANO model.}
    \label{fig:muscle_mapping}
\end{figure*}

\subsection{Joint-centric Muscle Adaptation}\label{sec:muscle_mapping}
In the physical human body, joints consist of either a single bone or a combination of multiple bones. Take the wrist joint, for example; it comprises \textit{the distal ends of the radius and ulna bones, 8 carpal bones, and the proximal segments of the 5 metacarpal bones} (\cref{fig:mapping-a}). The mappings between the joints and the bones are defined by academic consensus on anatomy.
In OpenSim, muscle data is documented based on its relative position to the bone it's attached to. To incorporate this muscle data into a joint-centric skinned model like MANO, we need to establish a mapping between muscles and joints. A direct method to achieve this is by using the existing "muscle-to-bone" relationships to formulate the "bone-to-joint" connections.

\paragraph{Muscle-to-Joint Mapping} We first analyze the MyoHand model and establish joint name mapping to the MANO model. Then, we transfer the origin point and insertion point of each muscle segment from bone-centric (\ie relative position to a certain bone) to joint-centric (\ie relative position to a certain joint). Specifically, let's denote the set of joints in the MANO model as $\mathcal{M} = \{m_i\}_{i=1}^n$, and the set of bone subgroups in the MyoHand model as $\mathcal{O} = \{O_i=\{b_{i,j}\}_{j=1}^k \}_{i=1}^n$, where $n$ is the number of joints, $\{b_j\}_{j=1}^k$ are the bones in the MyoHand model.
This mapping relationship can be represented as a function $f_\text{mapping}: O_i \mapsto m_j$.

The location at which the muscle-tendon connects is calculated by taking into account the relative position between the muscle tendon and the geometric mean center of the bone subgroup (which is regarded as the equivalent location of the MANO joint $m_j$).

For example, consider a point of attachment $q$ which has a displacement relative to a MyoHand bone $b_k$ expressed as $\text{dist}(q, b_k) \in \mathbb{R}^3$, the displacement $\text{dist}(q, m_j)$ relative to a MANO joint $m_j$ after the mapping can be calculated as
\begin{align}
    \bm{m}_j            & = \frac{\sum_{o \in O_j} \bm{o}}{|O_j|},       \\
    \text{dist}(q, m_j) & = \bm{q} - \bm{m_j}                            \\
                        & = (\bm{q} - \bm{b}_k) + (\bm{b}_k - \bm{m}_j)  \\
                        & = \text{dist}(q, b_k) + \text{dist}(b_k, m_j),
\end{align}
where $O_j = f^{-1}(m_j)$ represents the subgroup of bones associated with the joint $m_j$, and the bold symbols represent the location vectors of the bones and points.

Switching from a bone-focused to a joint-focused muscle representation immediately allows the muscle segment to connect solely to the joints. Thus, if the shape changes and alters the joint location, the muscle segment will adjust accordingly, as illustrated in \cref{fig:mapping-b}.

\paragraph{Manual Revision}

The MyoHand skeleton size may not align perfectly with MANO shapes due to their different human body sources, leading to issues such as muscle tendons intersecting the skin.  To address this, we collaborated with human experts to adjust the insertion points slightly. These adjustments ensure anatomical accuracy and compatibility with MyoHand's motion patterns, as shown in \cref{fig:mapping-c}.

\paragraph{Discussion}
As previously mentioned, the hand's musculoskeletal system is part of the entire upper extremity, with many muscle tendons starting in the forearm and ending in the hand. To incorporate this system into the MANO model, we integrated it with the SMPLX human body model. As shown in Figure \ref{fig:muscle_mapping}, our MS-MANO model includes the wrist and forearm. However, for consistency with existing datasets, we only focus on visualizing the hand in later experiments.

\begin{figure*}
    \centering
    \includegraphics[width=0.99\linewidth]{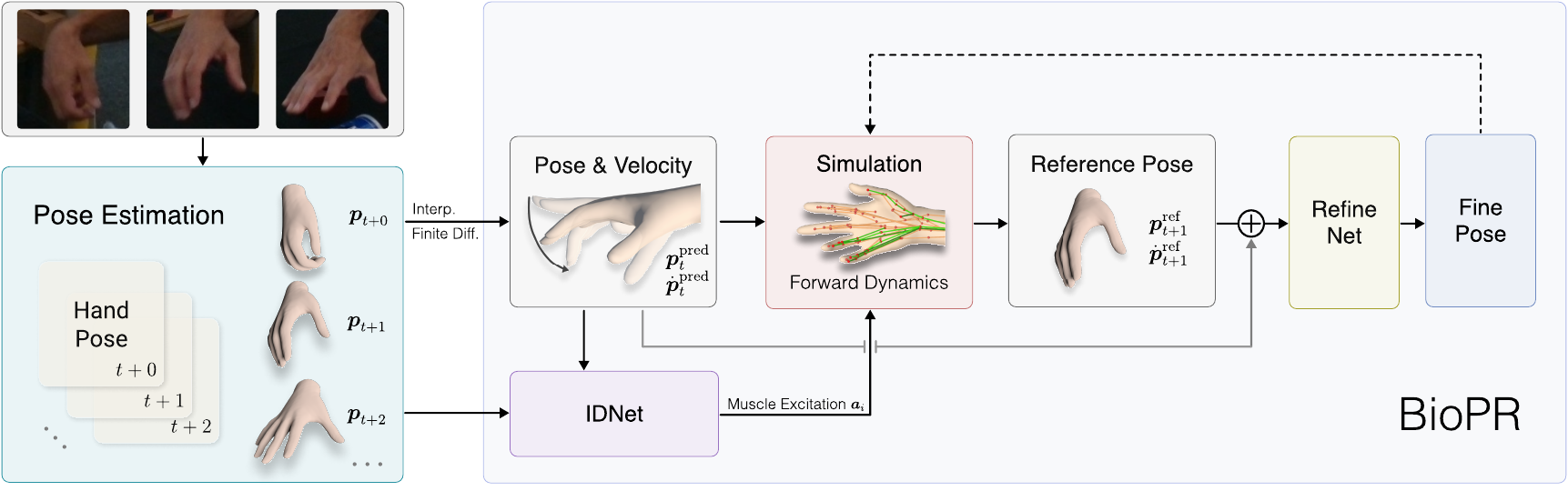}
    \caption{The simulation-in-the-loop pipeline of BioPR. Given a sequence of RGB images and the corresponding predictions of an existing hand pose estimator, BioPR first interpolates and differentiates the poses to get the joint velocities. Then, the IDNet is used to infer the muscle excitation signals. The joint poses, velocities, excitation signals, and the poses of the previous frame (denoted by dotted lines) are sent into the simulator, which will generate the next reference pose by forward kinematics. The Refine Net will do the final refinement based on the pose, velocity, and reference pose. On the next frame, the refined pose can be fed back to the simulator.}
    \label{fig:pipeline}
\end{figure*}

\section{Biomechanical Pose Refinement Framework, \textbf{BioPR}}

For the task of hand pose tracking, we begin with a video sequence $ \mathcal V = \{I_i\}_{i=1}^t$, which contains a single hand's movements. An off-the-shelf hand pose estimation algorithm is then applied to extract the predicted poses $ \mathcal{P}^{\text{pred}} = \{ \bm{p}^{\text{pred}}_i \} _{i=1}^t $ as well as the shape parameters for the MANO model.
Subsequently, these predicted poses are interpolated to calculate the velocity of the hand joints at a given time $t$, represented as $ \bm{v}_t $. By taking the observations near time $t$, we estimate the excitation signals $\bm a_i $ with an inverse dynamics network, IDNet (\cref{sec:inverse_dynamics}).
We run a forward dynamics simulation with the estimated excitation signals $\bm a$ to get a reference pose $\bm{p}^{\text{ref}}_i$ and velocity $\bm{v}^{\text{ref}}_i$.
These reference poses and velocities can create valid trajectories, which are beneficial for tasks such as motion generation.

However, for a visual analysis task in this work, due to the natural cross-individual differences in body shape and muscle structure, we cannot rely solely on simulated poses generated from standard muscle parameters. Therefore, we treat the reference pose as a biomechanical constraint and use a small neural network to produce a refined pose in a ``simulation-in-the-loop'' pipeline (\cref{sec:simulation_in_the_loop}).

\subsection{Muscle Inverse Dynamics and IDNet}\label{sec:inverse_dynamics}
To get the reference pose and velocity within biomechanical constraints, we first need to perform an inverse dynamics process to get the excitation signals $\bm a_i$, followed by a forward dynamics process. However, applying the inverse dynamics process to muscle is difficult.

Given a sequence of joint movements represented in axis-angle form as $\bm p_i \in \mathbb R^{n_{\text{joint}} \times 3}$ along with its angular velocity $\dot{\bm{p}}_i = \bm v_i\in \mathbb R^{n_{\text{joint}} \times 3 }$,
as well as the muscle excitation signal $\bm a_i \in \mathbb R^{n_{\text{muscle}}}$ that initiated the motion,
we define the inverse dynamics modeling for muscle actuators as:
\begin{equation}
    f_\text{inv}(\bm p_{i}, \bm v_{i}, \bm p_{i+1}, \bm v_{i+1}) = \bm a_i,
    \label{eq:finv}
\end{equation}
and the corresponding forward dynamics as:
\begin{equation}
    f_\text{fwd}(\bm p_i, \bm v_i, \bm a_i) = (\bm p_{i+1}, \bm v_{i+1}).
\end{equation}

In practice, the forward dynamics are computed using a physics engine to accurately simulate the physical interactions and constraints of the system. However, formulating inverse dynamics analytically is challenging, so we use a neural network, \textbf{IDNet}, to learn these dynamics.

Owing to the complexity of the human body, it is impossible to obtain accurate ground truth data of muscle excitation signals for each muscle.
We instead indirectly supervise these signals by comparing torques. The IDNet produces excitation signals $\bm a_i$, which are then used to compute the torque for each muscle, $\bm\tau_\text{m}$, as detailed in \cref{eq:torque}. For comparison, the reference torque (treated as ground truth) is calculated using a Proportional-Derivative (PD) controller with inverse dynamics compensation:
\begin{align}
    \bm\tau_\text{pd} & = k_p ({\bm p_d - \bm p})+ k_d ( \dot{\bm p}_d - \dot{\bm p} ),
\end{align}
where $\bm p_d, \dot{\bm p}_d$ are the desired pose and velocity.

We adopt a reinforcement learning algorithm, PPO~\cite{ppo}, to train the IDNet. The reward function is defined as
$$
    r = \exp\left( \omega_{\tau} \cdot \lVert \bm\tau_\text{pd} - \bm\tau_\text{m} \rVert\right),
$$
where $\omega_{\tau}$ is a constant.

The PD controller is only used to supervise training. In the testing process, the PD controller is disabled. During the testing phase, the PD controller is deactivated, and the muscle actuators are responsible for driving the motions.

\subsection{Simulation in the Loop}\label{sec:simulation_in_the_loop}

We have developed an approach called \textit{simulation-in-the-loop} to track hand movements consistently and accurately through a dynamic, iterative process, instead of relying on static estimates. This method simulates the internal dynamics of hand movements by considering the constraints imposed by the musculoskeletal system's biomechanics.

It initiates with the generation of initial hand pose estimates from video sequences using a base pose estimation algorithm. These algorithms are typically MANO-model-based and can produce joint and shape parameters.

Subsequently, we use the finite difference method to compute the derivatives of these estimated poses and result in the angular velocities $ \bm v^\text{pred} = \dot{\bm p}^\text{pred} \in \mathbb R^{15 \times 3}$ (the pose and velocity on the wrist joint are ignored). Our pipeline then processes pairs of consecutive pose and velocity data, $(\bm p_i, \bm v_i)$ and $(\bm p_{i+1}, \bm v_{i+1})$, through the IDNet $f_\text{inv}$ in \cref{eq:finv} to infer the muscle excitation signals $\bm a_i$ that facilitate pose transitions.

At each timestep, the simulator is updated with the current predicted position $\bm p^\text{pred}_i$ and velocity $\bm v^\text{pred}_i$. The inferred muscle excitation signals are then applied to simulate the next pose. As a result, we obtain a reference pose $\bm p^\text{ref}_{i+1}$, which adheres to the constraints of human anatomy and represents what the pose should plausibly be at the next timestep.

In the final stage, both the predicted pose $\bm p^\text{pred}_{i+1}$ and the reference pose $\bm p^\text{ref}_{i+1}$ are input into a refinement network, a multi-layer perceptron (MLP), which outputs a refined pose estimate, denoted by
\begin{equation}
    \bm p^\text{refined} = \mathcal{M}(\bm p^\text{pred}_{t+1}, \bm p^\text{ref}_{t+1}).
\end{equation}
The loss is defined by comparing with ground truth pose $\bm p_\text{gt}$
\begin{equation}
    \mathcal L_\text{refine} = \left\lVert \bm p^\text{gt} - \bm p^\text{refined} \right\rVert.
\end{equation}

\section{Experimental Setting}
\subsection{Datasets}
\paragraph{DexYCB}
The DexYCB dataset \cite{dexycb} is a large-scale dataset of hand-grasping postures captured using a synchronized setup of 8 cameras. It contains 20 common hand-held objects and 582K annotated 3D hand poses. For our experiments, we use the default \texttt{S0} training and testing split provided by the dataset, which separates the data by sequences.
The aligned viewpoints and the presence of occlusions of the DexYCB dataset present challenges to evaluating the robustness of our approach.

\paragraph{OakInk} The OakInk dataset \cite{oakink} is a large dataset for understanding hand-object interaction. It has 100 objects and 230k frames of hand poses, captured by a 4-camera setup. The dynamic forces exerted by objects on hands, due to the complex interaction sequences, challenge the stability and accuracy of our interaction simulations. In our experiments, we employ the default \texttt{SP0} split of the dataset, which splits the data by camera views.

\subsection{Metrics}
We use three different metrics to validate the proposed method.
The Mean Per Joint Position Error (\textbf{MPJPE}) in millimeters is the standard metric for hand pose estimation. It measures the mean joint distance error relative to the hand wrist over all 21 joints.
The Area Under the Curve (\textbf{AUC}) scores are provided by the official evaluation system of each dataset, which measures the robustness and precision across varying levels of joint thresholds.
The Acceleration Error (\textbf{AE}) in $\SI{}{\mm/\s^2}$ is used as a temporal consistency metric following previous works~\cite{deformer,vibe}.

\subsection{Simulation Setup}
Our simulation framework is based on the \texttt{RFUniverse} platform \cite{rfuniverse}. The hill-type muscle model is implemented based on the \texttt{Kinesis} package in Unity Engine. The human hand is controlled by 31 muscles responsible for facilitating movement beneath the wrist joint, and 8 muscles for controlling the wrist movements. The parameters for maximum isometric force and the length of the contractile element for each muscle are sourced from established MyoHand model data. To accurately capture the interactive dynamics involving the musculoskeletal system, our model incorporates anatomically aligned colliders that conform to the contours of human skeletal structures. Additionally, we introduce a minimal clearance around these colliders to effectively represent the deformable characteristics of human skin during collision events.

\subsection{Training Details}
\paragraph{IDNet} The IDNet is trained using Proximal Policy Optimization (PPO) \cite{ppo}, a common on-policy reinforcement learning method. Its input size is $16 \times 3 \times 4$ and output size is $31$, with two 256-d hidden layers. The network is trained on a NVIDIA A40 GPU. We use \texttt{RFUniverse} for pose, velocity, and muscle forward dynamics control. The training and simulation pipeline is vectorized. To be specific, we run 128 distributed processes on a platform with 2 AMD EPYC 7763 64-core processors. Each process controls 64 agents.
A small Gaussian noise $N(0, 0.1)$ in degree is applied to the joint rotations during the training process.

At each training step, we collect two consecutive frames from the simulator. Therefore, the total batch size is $128 \times 64 \times 2 = 16384$. The learning rate is $3 \times 10^{-4}$ with an adaptive scheduler \cite{stable-baselines3}. Each process runs at around 100 FPS, so we are able to generate the simulation data at around 10K FPS. It takes approximately 1 hour to train the IDNet.

\paragraph{Refine Net} Refine Net employs a Multi-Layer Perceptron (MLP) architecture and is trained sequentially following IDNet's convergence.
Its input size is $48 + 45 = 93$ (48 for the $\bm p^\text{pred}$ and 45 for the wrist-ignored $\bm p^\text{refined}$), and the output size is 48. The network only has a single 64-d hidden layer.
We use a learning rate of $1 \times 10^{-3}$ and a batch size of 10240 for training. We train the network for 4,500 iterations on a single NVIDIA A40 GPU, and it takes about 5 minutes.
\section{Results}
We evaluate the anatomic accuracy of the MS-MANO model in Sec. \ref{sec:mano_accuracy} and the efficacy of BioPR in Sec. \ref{sec:baselines_dexycb}.

\begin{figure*}
    \centering
    \includegraphics[width=1\linewidth]{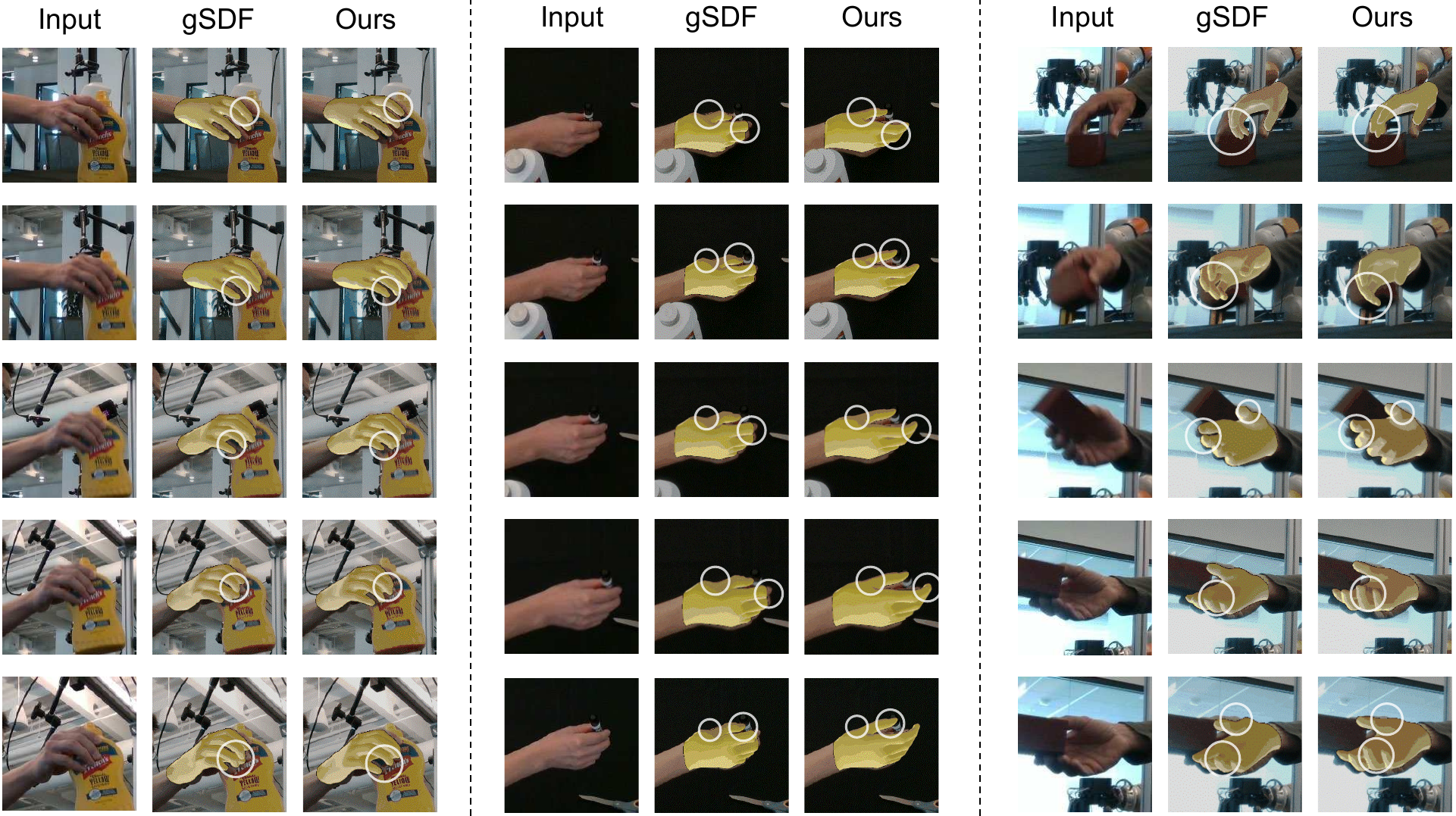}
    \caption{Qualitative results on DexYCB. \textbf{Left}: When a person is forcefully grasping a mustard bottle, there is a difference in the tightness of the middle, ring, and little fingers, comparing gSDF to our method. The projected results of our method better align with the input image. \textbf{Middle}: The thumb posture predicted by the gSDF method exhibits some odd distortion, which is not observed in our approach. \textbf{Right}: When there is severe occlusion, gSDF may generate some hand poses that lead to punctuation with the object. Our method mitigates such problem by catching the dynamics of the musculoskeletal system.}
    \label{fig:qual}
\end{figure*}

\begin{figure*}
    \centering
    \begin{subfigure}[b]{.32\linewidth}
        \centering
        \includegraphics[width=0.9\linewidth]{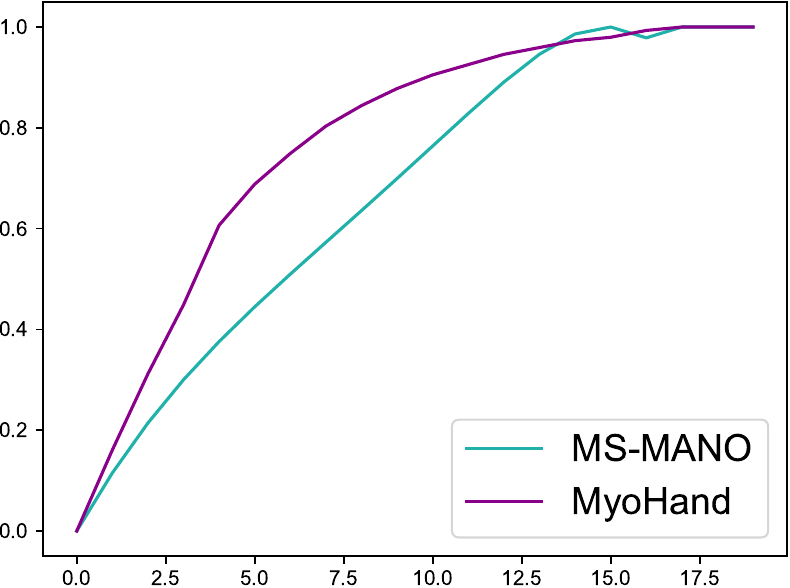}
        \caption{Joint Ring 1 Movements}
    \end{subfigure}
    \hfill
    \begin{subfigure}[b]{.32\linewidth}
        \centering
        \includegraphics[width=0.9\linewidth]{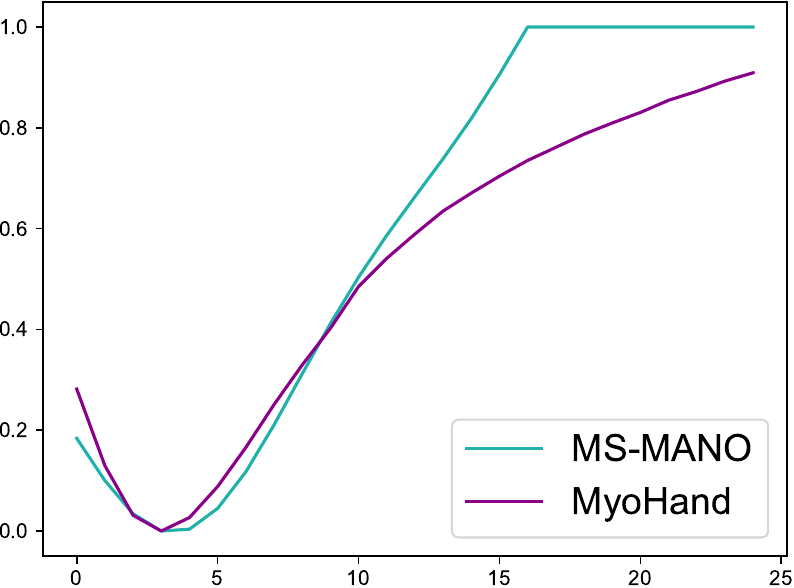}
        \caption{Joint Ring 2 Movements}
    \end{subfigure}
    \hfill
    \begin{subfigure}[b]{.32\linewidth}
        \centering
        \includegraphics[width=0.9\linewidth]{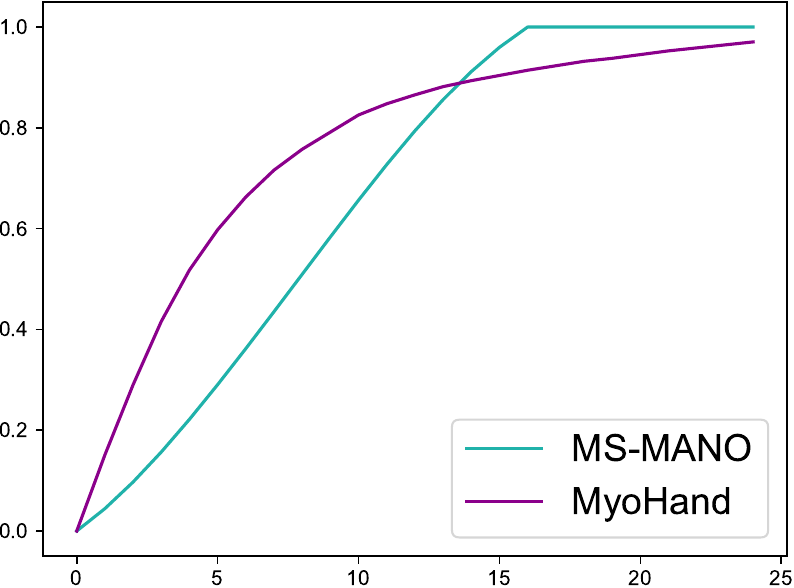}
        \caption{Joint Ring 3 Movements}
    \end{subfigure}
    \caption{The accuracy of simulation. The figures show normalized joint movements when exciting the FDS4\_R muscle, which can drive the ring finger to bend. The $x$ axis is the frames, and the $y$ axis is the relative joint movement.}
    \label{fig:simulation_acc}
\end{figure*}

\subsection{The Anatomic Accuracy of MS-MANO model}\label{sec:mano_accuracy}
To validate the anatomic accuracy of the proposed MS-MANO, we consulted anatomical experts in local hospitals and compared it with MyoHand \cite{myosuite} as it is a comprehensive musculoskeletal hand model built upon genuine anatomic data \cite{mobl1,mobl2,2ndhand}.
Figure \cref{fig:simulation_acc} presents the time-position plots for the ring finger bending trajectory in the MyoHand and MS-MANO models, monitoring three joints. The graphs indicate similar movement patterns in both models, with our model exhibiting a quicker response to muscle stimulation. This implies that our model activates faster, mirroring the immediate response of human muscles to neural signals, even at the extremes of joint motion.

\subsection{Comparison with Baselines on DexYCB}\label{sec:baselines_dexycb}

We compare the proposed method with some state-of-the-art methods.
Quantitative results on DexYCB are reported in \cref{tab:baselines_dexycb}. It shows that our method outperforms previous baseline methods. BioPR consistently and effectively enhances the base hand pose estimation method by refining the prediction results with the biomechanical constraints. Notably, the 3-layer IDNet and the 2-layer RefineNet have only 0.1M and 0.01M parameters, respectively. The BioPR framework only adds a minimal computational load to the base models, with an inference cost of just 9 ms.

\begin{table}
    \centering
    \begin{tabular}{c|ccc}
        \toprule
        Methods                              & MPJPE$\downarrow$ & AUC$\uparrow$ & AE$\downarrow$ \\ \midrule
        VIBE~\cite{vibe}                     & 16.95             & 67.5          & 36.4           \\
        TCMR~\cite{tcmr}                     & 16.03             & 70.1          & 34.3           \\
        MeshGraphormer~\cite{meshgraphormer} & 16.21             & 69.1          & 35.9           \\ \midrule
        gSDF~\cite{gsdf}                     & 14.4              & 89.1          & 30.3           \\
        gSDF + BioPR                         & \textbf{12.81}    & \textbf{89.7} & \textbf{29.9}  \\ \midrule
        Deformer~\cite{deformer}             & 13.64             & 89.6          & 31.7           \\
        Deformer + BioPR                     & \textbf{12.92}    & \textbf{90.4} & \textbf{30.7}  \\
        \bottomrule
    \end{tabular}
    \caption{Quantitative results on DexYCB.}
    \label{tab:baselines_dexycb}
\end{table}

Qualitative results are illustrated in \cref{fig:qual}, which shows the stability of the bio correction. We visualize our results by projecting the predicted 3D meshes onto the input images. Our approach demonstrates enhanced robustness in predicting poses that are aligned with anatomical structures. In contrast, previous methods exhibit certain artifacts, particularly unnatural twisting of thumb joints.

\subsection{Failure Cases}
\paragraph{Incorrect Annotations} Some of the failures come from the incorrect annotations of the ground truth. Since the annotations do not properly match the visual representation of the hand in the image,  training our method on the incorrect ground truth data leads to a misaligned result.

\paragraph{Errors from the Base Models} Occlusion, lighting, and various other factors can cause the base models to inaccurately predict results from the provided images. These inaccuracies can be persistent, enduring throughout the whole sequence. Our BioPR has the ability to correct it to some degree, but it still depends on the initial predictions from the base models to produce valid results.

\paragraph{Extremely Slow Motion}
A significant number of hand motion sequences exhibit minimal temporal variations. In these cases, the muscles are only slightly stimulated and generate small torques. These minor excitation signals fail to produce sufficient movement to achieve accurate frame-to-frame offsets, making the dynamics estimation and analysis challenging.

\subsection{Ablation Study}

\paragraph{Heuristics-based Priors}
To improve pose estimation in videos, where analyzing every frame is often unnecessary, we sample frames at intervals. This approach might lead to inconsistent motion predictions. We explored temporal smoothing, PCA, and TOCH \cite{track6} to mitigate this. Temporal smoothing averages poses within a window size $d$. The smoothed pose $\hat{\bm p}_i$ for a frame pose $\bm p_i \in I$ is determined by:
\begin{align}
    \hat{\bm p}_i
     & = \argmin\limits_{\bm m}\sum_{j=i-d/2}^{i+d/2}{\left\lVert\bm p_j - \bm m\right\rVert_2^2}.
\end{align}

Temporal smoothing and TOCH do not outperform our method, showing less than a 0.8mm improvement in MPJPE using gSDF with DexYCB dataset. PCA, on the other hand, led to a 7.91mm increase in MPJPE. These manually designed methods may not fully capture the complex dynamics due to the high dexterity of human hands.

However, our research shows that smoothing, while not significantly improving pose accuracy, creates realistic movement speeds. This outcome reduces sudden changes in muscle stimulation, resulting in smoother motion paths.

\paragraph{Muscle Insertion Points} The muscles are highly sensitive to their insertion points \cite{Ma2023.10.05.560588}.  Even minor alterations in these attachment points can lead to significant shifts in the trajectory of joint movement. We have experimented with various configurations of these insertion points and compared the trajectories before and after manual revision. The detailed visualization is presented in the supplementary materials. Such variations can cause the \textbf{IDNet} to struggle with convergence, at the same time resulting in noticeable artifacts within the joint movements.
\section{Conclusion}

In conclusion, our study presents an innovative approach to enhance hand dynamics analysis by integrating the musculoskeletal system into the learnable parametric hand model, MANO, resulting in the MS-MANO model. This model allows for movements that are more human-like and physiologically realistic and bridges the gap between image observations to biomechanics.
The BioPR refiner further refines hand pose estimations. Despite the challenges in creating the musculoskeletal model introduced by the complexity of the hand's dynamics system, our work validates the muscle adaptation by comparing the generated joint torques with MyoSuite. Moreover, the MS-MANO model and BioPR's effectiveness are evaluated on two large-scale public datasets, DexYCB and OakInk. The results showed consistent improvement in both quantitative and qualitative measures. Therefore, MS-MANO and BioPR mark a significant advancement in visual hand dynamics analysis, opening new avenues for future research and applications.
\section*{Acknowledgements}

We thank Yanbing Zhou for coding and figure support, Chen Bao for designing the initial codebase, Danqing Li and Yongkai Fan for muscle data processing, Yuan Fang and Haoyuan Fu for experimental help, and Jieyi Zhang, Wenxin Du and Han Xue for their insightful discussions.

This work was supported by 
the National Key Research and Development Project of China (No.~2022ZD0160102),
National Key Research and Development Project of China (No.~2021ZD0110704),
Shanghai Artificial Intelligence Laboratory,
XPLORER PRIZE grants.


{
    \small
    \bibliographystyle{ieeenat_fullname}
    \bibliography{main}

\begin{thebibliography}{53}
\providecommand{\natexlab}[1]{#1}
\providecommand{\url}[1]{\texttt{#1}}
\expandafter\ifx\csname urlstyle\endcsname\relax
  \providecommand{\doi}[1]{doi: #1}\else
  \providecommand{\doi}{doi: \begingroup \urlstyle{rm}\Url}\fi

\bibitem[Baek et~al.(2019)Baek, Kim, and Kim]{pose1}
Seungryul Baek, Kwang~In Kim, and Tae-Kyun Kim.
\newblock Pushing the envelope for rgb-based dense 3d hand pose estimation via neural rendering.
\newblock In \emph{Proceedings of the IEEE/CVF Conference on Computer Vision and Pattern Recognition}, pages 1067--1076, 2019.

\bibitem[Caggiano et~al.(2022)Caggiano, Wang, Durandau, Sartori, and Kumar]{myosuite}
Vittorio Caggiano, Huawei Wang, Guillaume Durandau, Massimo Sartori, and Vikash Kumar.
\newblock Myosuite--a contact-rich simulation suite for musculoskeletal motor control.
\newblock \emph{arXiv preprint arXiv:2205.13600}, 2022.

\bibitem[Cai et~al.(2018)Cai, Ge, Cai, and Yuan]{Cai_2018_ECCV}
Yujun Cai, Liuhao Ge, Jianfei Cai, and Junsong Yuan.
\newblock Weakly-supervised 3d hand pose estimation from monocular rgb images.
\newblock In \emph{Proceedings of the European Conference on Computer Vision (ECCV)}, 2018.

\bibitem[Cai et~al.(2019)Cai, Ge, Liu, Cai, Cham, Yuan, and Thalmann]{graph_track}
Yujun Cai, Liuhao Ge, Jun Liu, Jianfei Cai, Tat-Jen Cham, Junsong Yuan, and Nadia~Magnenat Thalmann.
\newblock Exploiting spatial-temporal relationships for 3d pose estimation via graph convolutional networks.
\newblock In \emph{Proceedings of the IEEE/CVF international conference on computer vision}, pages 2272--2281, 2019.

\bibitem[Chao et~al.(2021)Chao, Yang, Xiang, Molchanov, Handa, Tremblay, Narang, Van~Wyk, Iqbal, Birchfield, et~al.]{dexycb}
Yu-Wei Chao, Wei Yang, Yu Xiang, Pavlo Molchanov, Ankur Handa, Jonathan Tremblay, Yashraj~S Narang, Karl Van~Wyk, Umar Iqbal, Stan Birchfield, et~al.
\newblock Dexycb: A benchmark for capturing hand grasping of objects.
\newblock In \emph{Proceedings of the IEEE/CVF Conference on Computer Vision and Pattern Recognition}, pages 9044--9053, 2021.

\bibitem[Chen et~al.(2023)Chen, Chen, Schmid, and Laptev]{gsdf}
Zerui Chen, Shizhe Chen, Cordelia Schmid, and Ivan Laptev.
\newblock gsdf: Geometry-driven signed distance functions for 3d hand-object reconstruction.
\newblock In \emph{Proceedings of the IEEE/CVF Conference on Computer Vision and Pattern Recognition}, pages 12890--12900, 2023.

\bibitem[Choi et~al.(2021)Choi, Moon, Chang, and Lee]{tcmr}
Hongsuk Choi, Gyeongsik Moon, Ju~Yong Chang, and Kyoung~Mu Lee.
\newblock Beyond static features for temporally consistent 3d human pose and shape from a video.
\newblock In \emph{Conference on Computer Vision and Pattern Recognition (CVPR)}, 2021.

\bibitem[Fan et~al.(2023)Fan, Taheri, Tzionas, Kocabas, Kaufmann, Black, and Hilliges]{arctic}
Zicong Fan, Omid Taheri, Dimitrios Tzionas, Muhammed Kocabas, Manuel Kaufmann, Michael~J Black, and Otmar Hilliges.
\newblock Arctic: A dataset for dexterous bimanual hand-object manipulation.
\newblock In \emph{Proceedings of the IEEE/CVF Conference on Computer Vision and Pattern Recognition}, pages 12943--12954, 2023.

\bibitem[Fu et~al.()Fu, Xu, Ye, Xue, Yu, Tang, Li, Du, Zhang, and Lu]{rfuniverse}
Haoyuan Fu, Wenqiang Xu, Ruolin Ye, Han Xue, Zhenjun Yu, Tutian Tang, Yutong Li, Wenxin Du, Jieyi Zhang, and Cewu Lu.
\newblock Demonstrating rfuniverse: A multiphysics simulation platform for embodied ai.

\bibitem[Fu et~al.(2023)Fu, Liu, Xu, Niebles, and Kitani]{deformer}
Qichen Fu, Xingyu Liu, Ran Xu, Juan~Carlos Niebles, and Kris~M Kitani.
\newblock Deformer: Dynamic fusion transformer for robust hand pose estimation.
\newblock \emph{arXiv preprint arXiv:2303.04991}, 2023.

\bibitem[Geijtenbeek et~al.(2013)Geijtenbeek, Van De~Panne, and Van Der~Stappen]{locomotion1}
Thomas Geijtenbeek, Michiel Van De~Panne, and A~Frank Van Der~Stappen.
\newblock Flexible muscle-based locomotion for bipedal creatures.
\newblock \emph{ACM Transactions on Graphics (TOG)}, 32\penalty0 (6):\penalty0 1--11, 2013.

\bibitem[Hasson et~al.(2020)Hasson, Tekin, Bogo, Laptev, Pollefeys, and Schmid]{track1}
Yana Hasson, Bugra Tekin, Federica Bogo, Ivan Laptev, Marc Pollefeys, and Cordelia Schmid.
\newblock Leveraging photometric consistency over time for sparsely supervised hand-object reconstruction.
\newblock In \emph{Proceedings of the IEEE/CVF conference on computer vision and pattern recognition}, pages 571--580, 2020.

\bibitem[Hill(1938)]{hilltype}
Archibald~Vivian Hill.
\newblock The heat of shortening and the dynamic constants of muscle.
\newblock \emph{Proceedings of the Royal Society of London. Series B-Biological Sciences}, 1938.

\bibitem[Jiang et~al.(2019)Jiang, Van~Wouwe, De~Groote, and Liu]{muscle_joint}
Yifeng Jiang, Tom Van~Wouwe, Friedl De~Groote, and C~Karen Liu.
\newblock Synthesis of biologically realistic human motion using joint torque actuation.
\newblock \emph{ACM Transactions On Graphics (TOG)}, 38\penalty0 (4):\penalty0 1--12, 2019.

\bibitem[Kocabas et~al.(2020{\natexlab{a}})Kocabas, Athanasiou, and Black]{track2}
Muhammed Kocabas, Nikos Athanasiou, and Michael~J Black.
\newblock Vibe: Video inference for human body pose and shape estimation.
\newblock In \emph{Proceedings of the IEEE/CVF conference on computer vision and pattern recognition}, pages 5253--5263, 2020{\natexlab{a}}.

\bibitem[Kocabas et~al.(2020{\natexlab{b}})Kocabas, Athanasiou, and Black]{vibe}
Muhammed Kocabas, Nikos Athanasiou, and Michael~J. Black.
\newblock Vibe: Video inference for human body pose and shape estimation.
\newblock In \emph{The IEEE Conference on Computer Vision and Pattern Recognition (CVPR)}, 2020{\natexlab{b}}.

\bibitem[Komura et~al.(2000)Komura, Shinagawa, and Kunii]{create_retarget}
Taku Komura, Yoshihisa Shinagawa, and Tosiyasu~L Kunii.
\newblock Creating and retargetting motion by the musculoskeletal human body model.
\newblock \emph{The visual computer}, 16:\penalty0 254--270, 2000.

\bibitem[Lee et~al.(2015)Lee, Asakawa, Dennerlein, and Jindrich]{2ndhand}
Jong~Hwa Lee, Deanna~S Asakawa, Jack~T Dennerlein, and Devin~L Jindrich.
\newblock Finger muscle attachments for an opensim upper-extremity model.
\newblock \emph{PloS one}, 10\penalty0 (4):\penalty0 e0121712, 2015.

\bibitem[Lee et~al.(2018)Lee, Yu, Park, Aanjaneya, Sifakis, and Lee]{upper1}
Seunghwan Lee, Ri Yu, Jungnam Park, Mridul Aanjaneya, Eftychios Sifakis, and Jehee Lee.
\newblock Dexterous manipulation and control with volumetric muscles.
\newblock \emph{ACM Transactions on Graphics (TOG)}, 37\penalty0 (4):\penalty0 1--13, 2018.

\bibitem[Lee et~al.(2019)Lee, Park, Lee, and Lee]{mass}
Seunghwan Lee, Moonseok Park, Kyoungmin Lee, and Jehee Lee.
\newblock Scalable muscle-actuated human simulation and control.
\newblock \emph{ACM Transactions On Graphics (TOG)}, 38\penalty0 (4):\penalty0 1--13, 2019.

\bibitem[Lee and Terzopoulos(2006)]{upper3}
Sung-Hee Lee and Demetri Terzopoulos.
\newblock Heads up! biomechanical modeling and neuromuscular control of the neck.
\newblock In \emph{ACM SIGGRAPH 2006 Papers}, pages 1188--1198. 2006.

\bibitem[Lee et~al.(2009)Lee, Sifakis, and Terzopoulos]{upper2}
Sung-Hee Lee, Eftychios Sifakis, and Demetri Terzopoulos.
\newblock Comprehensive biomechanical modeling and simulation of the upper body.
\newblock \emph{ACM Transactions on Graphics (TOG)}, 28\penalty0 (4):\penalty0 1--17, 2009.

\bibitem[Lee et~al.(2014)Lee, Park, Kwon, and Lee]{locomotion4}
Yoonsang Lee, Moon~Seok Park, Taesoo Kwon, and Jehee Lee.
\newblock Locomotion control for many-muscle humanoids.
\newblock \emph{ACM Transactions on Graphics (TOG)}, 33\penalty0 (6):\penalty0 1--11, 2014.

\bibitem[Lin et~al.(2021)Lin, Wang, and Liu]{meshgraphormer}
Kevin Lin, Lijuan Wang, and Zicheng Liu.
\newblock Mesh graphormer.
\newblock In \emph{ICCV}, 2021.

\bibitem[Liu et~al.(2021)Liu, Jiang, Xu, Liu, and Wang]{pose3}
Shaowei Liu, Hanwen Jiang, Jiarui Xu, Sifei Liu, and Xiaolong Wang.
\newblock Semi-supervised 3d hand-object poses estimation with interactions in time.
\newblock In \emph{Proceedings of the IEEE/CVF Conference on Computer Vision and Pattern Recognition}, pages 14687--14697, 2021.

\bibitem[Loper et~al.(2015)Loper, Mahmood, Romero, Pons-Moll, and Black]{smpl}
Matthew Loper, Naureen Mahmood, Javier Romero, Gerard Pons-Moll, and Michael~J. Black.
\newblock {SMPL}: A skinned multi-person linear model.
\newblock \emph{ACM TOG}, 34\penalty0 (6):\penalty0 248:1--248:16, 2015.

\bibitem[Lorensen and Cline(1987)]{marching_cubes}
William~E Lorensen and Harvey~E Cline.
\newblock Marching cubes: A high resolution 3d surface construction algorithm.
\newblock \emph{ACM TOG}, 21\penalty0 (4):\penalty0 163--169, 1987.

\bibitem[Lv et~al.(2021)Lv, Xu, Yang, Qian, Mao, and Lu]{handtailor}
Jun Lv, Wenqiang Xu, Lixin Yang, Sucheng Qian, Chongzhao Mao, and Cewu Lu.
\newblock Handtailor: Towards high-precision monocular 3d hand recovery.
\newblock 2021.

\bibitem[Ma et~al.(2023)Ma, Guerra, Caillet, Zhao, Clarke, Maksymenko, Deslauriers-Gauthier, Sheng, Zhu, and Farina]{Ma2023.10.05.560588}
Shihan Ma, Irene~Mendez Guerra, Arnault~Hubert Caillet, Jiamin Zhao, Alexander~Kenneth Clarke, Kostiantyn Maksymenko, Samuel Deslauriers-Gauthier, Xinjun Sheng, Xiangyang Zhu, and Dario Farina.
\newblock Neuromotion: Open-source simulator with neuromechanical and deep network models to generate surface emg signals during voluntary movement.
\newblock 2023.

\bibitem[McFarland et~al.(2019)McFarland, McCain, Poppo, and Saul]{mobl2}
Daniel~C McFarland, Emily~M McCain, Michael~N Poppo, and Katherine~R Saul.
\newblock Spatial dependency of glenohumeral joint stability during dynamic unimanual and bimanual pushing and pulling.
\newblock \emph{Journal of biomechanical engineering}, 141\penalty0 (5):\penalty0 051006, 2019.

\bibitem[Mordatch et~al.(2013)Mordatch, Wang, Todorov, and Koltun]{locomotion2}
Igor Mordatch, Jack~M Wang, Emanuel Todorov, and Vladlen Koltun.
\newblock Animating human lower limbs using contact-invariant optimization.
\newblock \emph{ACM Transactions on Graphics (TOG)}, 32\penalty0 (6):\penalty0 1--8, 2013.

\bibitem[Park et~al.(2022)Park, Oh, Moon, Choi, and Lee]{track3}
JoonKyu Park, Yeonguk Oh, Gyeongsik Moon, Hongsuk Choi, and Kyoung~Mu Lee.
\newblock Handoccnet: Occlusion-robust 3d hand mesh estimation network.
\newblock In \emph{Proceedings of the IEEE/CVF Conference on Computer Vision and Pattern Recognition}, pages 1496--1505, 2022.

\bibitem[Pavlakos et~al.(2019)Pavlakos, Choutas, Ghorbani, Bolkart, Osman, Tzionas, and Black]{smplx}
Georgios Pavlakos, Vasileios Choutas, Nima Ghorbani, Timo Bolkart, Ahmed~AA Osman, Dimitrios Tzionas, and Michael~J Black.
\newblock Expressive body capture: 3d hands, face, and body from a single image.
\newblock In \emph{CVPR}, 2019.

\bibitem[Raffin et~al.(2021)Raffin, Hill, Gleave, Kanervisto, Ernestus, and Dormann]{stable-baselines3}
Antonin Raffin, Ashley Hill, Adam Gleave, Anssi Kanervisto, Maximilian Ernestus, and Noah Dormann.
\newblock Stable-baselines3: Reliable reinforcement learning implementations.
\newblock \emph{Journal of Machine Learning Research}, 22\penalty0 (268):\penalty0 1--8, 2021.

\bibitem[Romero et~al.(2017)Romero, Tzionas, and Black]{mano}
Javier Romero, Dimitrios Tzionas, and Michael~J. Black.
\newblock Embodied hands: Modeling and capturing hands and bodies together.
\newblock \emph{ACM Transactions on Graphics, (Proc. SIGGRAPH Asia)}, 2017.

\bibitem[Saul et~al.(2015)Saul, Hu, Goehler, Vidt, Daly, Velisar, and Murray]{mobl1}
Katherine~R Saul, Xiao Hu, Craig~M Goehler, Meghan~E Vidt, Melissa Daly, Anca Velisar, and Wendy~M Murray.
\newblock Benchmarking of dynamic simulation predictions in two software platforms using an upper limb musculoskeletal model.
\newblock \emph{Computer methods in biomechanics and biomedical engineering}, 18\penalty0 (13):\penalty0 1445--1458, 2015.

\bibitem[Schleicher et~al.(2021)Schleicher, Nitschke, Martschinke, Stamminger, Eskofier, Klucken, and Koelewijn]{bash}
Robert Schleicher, Marlies Nitschke, Jana Martschinke, Marc Stamminger, Bj{\"o}rn~M Eskofier, Jochen Klucken, and Anne~D Koelewijn.
\newblock Bash: Biomechanical animated skinned human for visualization of kinematics and muscle activity.
\newblock In \emph{VISIGRAPP (1: GRAPP)}, pages 25--36, 2021.

\bibitem[Schulman et~al.(2017)Schulman, Wolski, Dhariwal, Radford, and Klimov]{ppo}
John Schulman, Filip Wolski, Prafulla Dhariwal, Alec Radford, and Oleg Klimov.
\newblock Proximal policy optimization algorithms.
\newblock \emph{CoRR}, abs/1707.06347, 2017.

\bibitem[Seth et~al.(2018)Seth, Hicks, Uchida, Habib, Dembia, Dunne, Ong, DeMers, Rajagopal, Millard, et~al.]{opensim}
Ajay Seth, Jennifer~L Hicks, Thomas~K Uchida, Ayman Habib, Christopher~L Dembia, James~J Dunne, Carmichael~F Ong, Matthew~S DeMers, Apoorva Rajagopal, Matthew Millard, et~al.
\newblock Opensim: Simulating musculoskeletal dynamics and neuromuscular control to study human and animal movement.
\newblock \emph{PLoS computational biology}, 14\penalty0 (7):\penalty0 e1006223, 2018.

\bibitem[Si et~al.(2014)Si, Lee, Sifakis, and Terzopoulos]{locomotion3}
Weiguang Si, Sung-Hee Lee, Eftychios Sifakis, and Demetri Terzopoulos.
\newblock Realistic biomechanical simulation and control of human swimming.
\newblock \emph{ACM Transactions on Graphics (TOG)}, 34\penalty0 (1):\penalty0 1--15, 2014.

\bibitem[Sueda et~al.(2008)Sueda, Kaufman, and Pai]{hand_manip1}
Shinjiro Sueda, Andrew Kaufman, and Dinesh~K Pai.
\newblock Musculotendon simulation for hand animation.
\newblock In \emph{ACM SIGGRAPH 2008 papers}, pages 1--8. 2008.

\bibitem[Tiwari et~al.(2022)Tiwari, Anti{\'c}, Lenssen, Sarafianos, Tung, and Pons-Moll]{track4}
Garvita Tiwari, Dimitrije Anti{\'c}, Jan~Eric Lenssen, Nikolaos Sarafianos, Tony Tung, and Gerard Pons-Moll.
\newblock Pose-ndf: Modeling human pose manifolds with neural distance fields.
\newblock In \emph{European Conference on Computer Vision}, pages 572--589. Springer, 2022.

\bibitem[Tsang et~al.(2005)Tsang, Singh, and Fiume]{hand_manip2}
Winnie Tsang, Karan Singh, and Eugene Fiume.
\newblock Helping hand: an anatomically accurate inverse dynamics solution for unconstrained hand motion.
\newblock In \emph{Proceedings of the 2005 ACM SIGGRAPH/Eurographics symposium on Computer animation}, pages 319--328, 2005.

\bibitem[Wang et~al.(2012)Wang, Hamner, Delp, and Koltun]{muscle_energy}
Jack~M Wang, Samuel~R Hamner, Scott~L Delp, and Vladlen Koltun.
\newblock Optimizing locomotion controllers using biologically-based actuators and objectives.
\newblock \emph{ACM Transactions on Graphics (TOG)}, 31\penalty0 (4):\penalty0 1--11, 2012.

\bibitem[Yang et~al.(2020{\natexlab{a}})Yang, Chang, Lee, and Kwak]{track5}
John Yang, Hyung~Jin Chang, Seungeui Lee, and Nojun Kwak.
\newblock Seqhand: Rgb-sequence-based 3d hand pose and shape estimation.
\newblock In \emph{Computer Vision--ECCV 2020: 16th European Conference, Glasgow, UK, August 23--28, 2020, Proceedings, Part XII 16}, pages 122--139. Springer, 2020{\natexlab{a}}.

\bibitem[Yang et~al.(2020{\natexlab{b}})Yang, Li, Xu, Diao, and Lu]{bihand}
Lixin Yang, Jiasen Li, Wenqiang Xu, Yiqun Diao, and Cewu Lu.
\newblock Bihand: Recovering hand mesh with multi-stage bisected hourglass networks.
\newblock In \emph{{BMVC} British Machine Vision Conference}, 2020{\natexlab{b}}.

\bibitem[Yang et~al.(2021)Yang, Zhan, Li, Xu, Li, and Lu]{cpf2}
Lixin Yang, Xinyu Zhan, Kailin Li, Wenqiang Xu, Jiefeng Li, and Cewu Lu.
\newblock Cpf: Learning a contact potential field to model the hand-object interaction.
\newblock In \emph{Proceedings of the IEEE/CVF International Conference on Computer Vision}, pages 11097--11106, 2021.

\bibitem[Yang et~al.(2022{\natexlab{a}})Yang, Li, Zhan, Lv, Xu, Li, and Lu]{artiboost}
Lixin Yang, Kailin Li, Xinyu Zhan, Jun Lv, Wenqiang Xu, Jiefeng Li, and Cewu Lu.
\newblock Artiboost: Boosting articulated 3d hand-object pose estimation via online exploration and synthesis.
\newblock In \emph{{CVPR} IEEE Conference on Computer Vision and Pattern Recognition}, pages 2750--2760, 2022{\natexlab{a}}.

\bibitem[Yang et~al.(2022{\natexlab{b}})Yang, Li, Zhan, Wu, Xu, Liu, and Lu]{oakink}
Lixin Yang, Kailin Li, Xinyu Zhan, Fei Wu, Anran Xu, Liu Liu, and Cewu Lu.
\newblock Oakink: A large-scale knowledge repository for understanding hand-object interaction.
\newblock In \emph{Proceedings of the IEEE/CVF Conference on Computer Vision and Pattern Recognition}, pages 20953--20962, 2022{\natexlab{b}}.

\bibitem[Yang et~al.(2024)Yang, Zhan, Li, Xu, Zhang, Li, and Lu]{cpf1}
Lixin Yang, Xinyu Zhan, Kailin Li, Wenqiang Xu, Junming Zhang, Jiefeng Li, and Cewu Lu.
\newblock Learning a contact potential field for modeling the hand-object interaction.
\newblock \emph{IEEE transactions on pattern analysis and machine intelligence}, 2024.

\bibitem[Ye et~al.(2021)Ye, Xu, Xue, Tang, Wang, and Lu]{h2o}
Ruolin Ye, Wenqiang Xu, Zhendong Xue, Tutian Tang, Yanfeng Wang, and Cewu Lu.
\newblock H2o: A benchmark for visual human-human object handover analysis.
\newblock In \emph{{ICCV} IEEE/CVF International Conference on Computer Vision}, pages 15762--15771, 2021.

\bibitem[Ye et~al.(2022)Ye, Xu, Fu, Jenamani, Nguyen, Lu, Dimitropoulou, and Bhattacharjee]{rcare}
Ruolin Ye, Wenqiang Xu, Haoyuan Fu, Rajat~Kumar Jenamani, Vy Nguyen, Cewu Lu, Katherine Dimitropoulou, and Tapomayukh Bhattacharjee.
\newblock Rcare world: A human-centric simulation world for caregiving robots.
\newblock In \emph{2022 IEEE/RSJ International Conference on Intelligent Robots and Systems (IROS)}, pages 33--40. IEEE, 2022.

\bibitem[Zhou et~al.(2022)Zhou, Bhatnagar, Lenssen, and Pons-Moll]{track6}
Keyang Zhou, Bharat~Lal Bhatnagar, Jan~Eric Lenssen, and Gerard Pons-Moll.
\newblock Toch: Spatio-temporal object-to-hand correspondence for motion refinement.
\newblock In \emph{European Conference on Computer Vision}, pages 1--19. Springer, 2022.

\end{thebibliography}
}

\end{document}